\documentclass[letterpaper]{article} 
\usepackage[draft]{aaai25}  
\usepackage{times}  
\usepackage{helvet}  
\usepackage{courier}  
\usepackage[hyphens]{url}  
\usepackage{graphicx} 
\usepackage{natbib}  
\usepackage{caption} 
\usepackage{algorithm}
\usepackage{algpseudocode}
\usepackage{float}
\usepackage{lscape}
\usepackage{booktabs}
\usepackage{array}
\usepackage{amsmath}

\usepackage{newfloat}
\usepackage{listings}
\DeclareCaptionStyle{ruled}{labelfont=normalfont,labelsep=colon,strut=off} 
\floatstyle{ruled}
\newfloat{listing}{tb}{lst}{}
\floatname{listing}{Listing}
\title{AdaComp: Extractive Context Compression with Adaptive Predictor \\ for Retrieval-Augmented Large Language Models}

\author {
    Qianchi Zhang\textsuperscript{\rm 1,\rm 2},  
    Hainan Zhang\textsuperscript{\rm 1,\rm 2}\thanks{Corresponding author.}, 
    Liang Pang\textsuperscript{\rm 4}, 
    Hongwei Zheng\textsuperscript{\rm 3},
    Zhiming Zheng\textsuperscript{\rm 1,\rm 2}
}
\affiliations {
    \textsuperscript{\rm 1}Beijing Advanced Innovation Center for Future Blockchain and Privacy Computing \\
\textsuperscript{\rm 2}School of Artificial Intelligence, Beihang University \\
    \textsuperscript{\rm 3}Beijing Academy of Blockchain and Edge Computing \\
    \textsuperscript{\rm 4}Institute of Computing Technology, Chinese Academy of Sciences \\
    \texttt{\{zhangqianchi, zhanghainan\}@buaa.edu.cn}
}

\usepackage{bibentry}

\begin{document}

\maketitle

\begin{abstract}

Retrieved documents containing noise will hinder Retrieval-Augmented Generation (RAG) from detecting answer clues and make the inference process slow and expensive. Therefore, context compression is necessary to enhance its accuracy and efficiency. Existing context compression methods use extractive or generative models to retain the most query-relevant sentences or apply the information bottleneck theory to preserve sufficient information. However, these methods may face issues such as over-compression or high computational costs. 
We observe that the retriever often ranks relevant documents at the top, but the exact number of documents needed to answer the query is uncertain due to the impact of query complexity and retrieval quality: complex queries like multi-hop questions may require retaining more documents than simpler queries, and a low-quality retrieval may need to rely on more documents to generate accurate outputs. Therefore, determining the minimum number of required documents (compression rate) is still a challenge for RAG.
In this paper, we introduce AdaComp, a low-cost extractive context compression method that adaptively determines the compression rate based on both query complexity and retrieval quality. Specifically, we first annotate the minimum top-k documents necessary for the RAG system to answer the current query as the compression rate and then construct triplets of the query, retrieved documents, and its compression rate. Then, we use this triplet dataset to train a compression-rate predictor. During inference, the compressor adaptively selects the top-k documents as the context-filtering documents based on the predictor's output and performs LLM inference. 
Experiments on three QA datasets and one conversational Multi-doc QA dataset show that AdaComp significantly reduces inference costs while maintaining performance nearly identical to uncompressed models, achieving a balance between efficiency and performance.
\end{abstract}

\begin{figure}[!t]
\centering
\includegraphics[width=0.9\columnwidth]{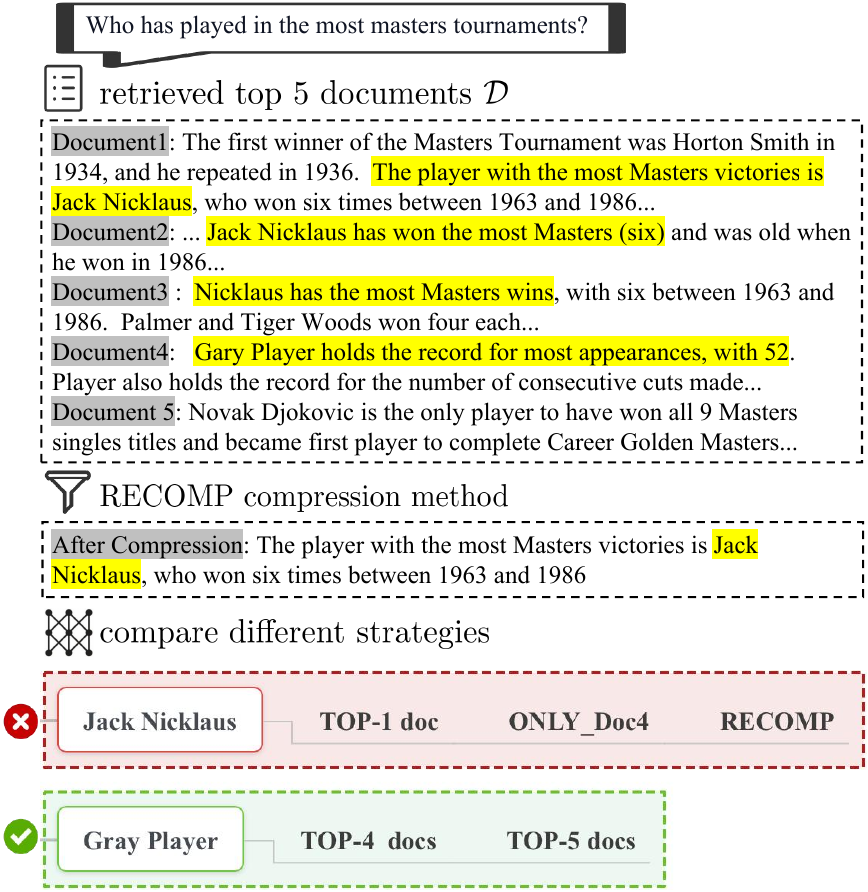} 
\caption{An illustration of how retrieval quality affects the generation results of context compression models. TOP-1 and RECOMP select the most query-relevant sentences, but they produce incorrect answers due to over-compression. ONLY\_Doc4 select the $4^{th}$ document as context but it can not answer correctly because it lacks background knowledge about the query. Although TOP-5 can answer correctly, document 5 is irrelevant and should be filtered out.}
\label{fig:figure1}
\end{figure}

\section{Introduction}

Retrieval-Augmented Generation (RAG) has demonstrated impressive performance across various knowledge-intensive NLP tasks, such as open-domain question answering~\cite{mao2021generation,sun2021covid,zhang2026less,zhang2026stable}, fact verification~\cite{chen2022gere}, and knowledge-grounded dialogue generation~\cite{huang2023learning}. 
It improves the relevance, coherence, and factual accuracy of outputs by appending a large number of retrieved documents to the query as context~\cite{gao2023retrieval}.
However, the effectiveness of the current RAG heavily depends on the relevance of retrieved documents~\cite{liu2024lost}.
When retrieved documents contain noise or irrelevant information, the generation model struggles to detect answer clues because noise interferes with self-attention's ability to reason over the correct context. Moreover, the inference will also become slow and costly~\cite{zhu2024information}. Therefore, it is crucial to filter out irrelevant and low-value contexts.

Current context compression methods mainly use extractive or generative models to compress the retrieved documents, but they may face issues of over-compression and high computational costs. \citet{xu2023recomp} introduce RECOMP to select the most query-relevant sentences as filtered context, but it may struggle with complex queries, such as multi-hop or open-ended questions, because it over-compresses the context, leading to a decline in RAG performance. The Information Bottleneck Theory~\cite{zhu2024information} tries to use reinforcement learning to find the optimal compression strategy by maximizing the mutual information between compressed data and the actual output while minimizing the mutual information between compressed data and retrieved documents. However, its high computational costs make it difficult to quickly adapt to various retrieval systems in the real world. Therefore, a more efficient and low-cost method is needed to retain sufficient context for RAG while minimizing noise and computational overhead.

We find that in most cases, the retriever can rank relevant documents at the top, but the exact number of documents needed to answer the query is uncertain\footnote{The reasons why we use truncation to calculate the compression rate are: (1) we evaluated RAG accuracy on three QA datasets using different top-k documents as context. The results (see Experiments~\ref{sec:moExam}) show that accuracy first increases and then decreases, indicating that finding the optimal truncation point can enhance RAG accuracy; (2) we also tried using only the k-th document as context, but its performance is worse than the top-1(see Table~\ref{tab:table1}), likely because the front documents provide essential background.} due to the impact of query complexity and retrieval quality. For example, complex multi-hop questions require more documents for comprehensive judgment, and open-ended questions need broader background knowledge to provide a well-rounded answer, such as ``How to take care of a little cat?''. What's more, retrieval quality can also influence the number of required documents. When the retrieval quality is high, even complex questions may be answered with just the most relevant information. Conversely, when it is poor, as shown in Figure~\ref{fig:figure1}, even simple questions may require synthesizing more information to arrive at the correct answer. Therefore, determining the number of minimum required documents(compression rate) should consider both the question and the retrieved documents, quickly adapting to find the optimal compression strategy for different retrievers.

In this paper, we propose a low-cost extractive context compression method, named AdaComp, which adaptively determines the compression rate based on both query complexity and retrieval quality. Specifically, we first annotate the minimum top-k documents required by the RAG system to accurately answer the query as the compression rate, and then construct triplets consisting of the query, retrieved documents, and compression rate. We then concatenate the query with retrieved documents to form the input and use the compression rate from the triplets as the output to train a compression-rate predictor. This predictor can adaptively determine the top-k documents needed by the current RAG system based on query complexity and the retrieved documents. During inference, the compressor selects the top-k documents as the filtered context based on the predictor’s output, ensuring the RAG system has access to concise and sufficient information. This method enables quick and low-cost adaptation to various RAG systems without the need for multiple inferences.

We conduct experiments on three open-domain question answering datasets, i.e., NQ, TriviaQA, HotpotQA, and one conversational Multi-doc QA dataset. The results show that AdaComp outperforms the baseline models in maintaining performance while significantly reducing the context required for inference. Further analysis of various difficulty queries demonstrates that AdaComp is more accurate in perceiving the required amount of documents, thereby validating the effectiveness of our approach.

To summarize, our main contributions are as follows: 
\begin{itemize}
    \item We propose an automated and low-cost compression rate annotation method that determines the minimum required top-k documents based on the system's real ability, enabling quick adaptation to various RAG systems.
    
    \item We design an effective extractive context compression method, which determines the compression rate based on both query complexity and retrieval quality.
    
    \item Experiments on four datasets show that filtering out unimportant noisy documents improves inference efficiency while maintaining performance.
\end{itemize}

\begin{figure*}[t]
\centering
\includegraphics[width=0.8\textwidth]{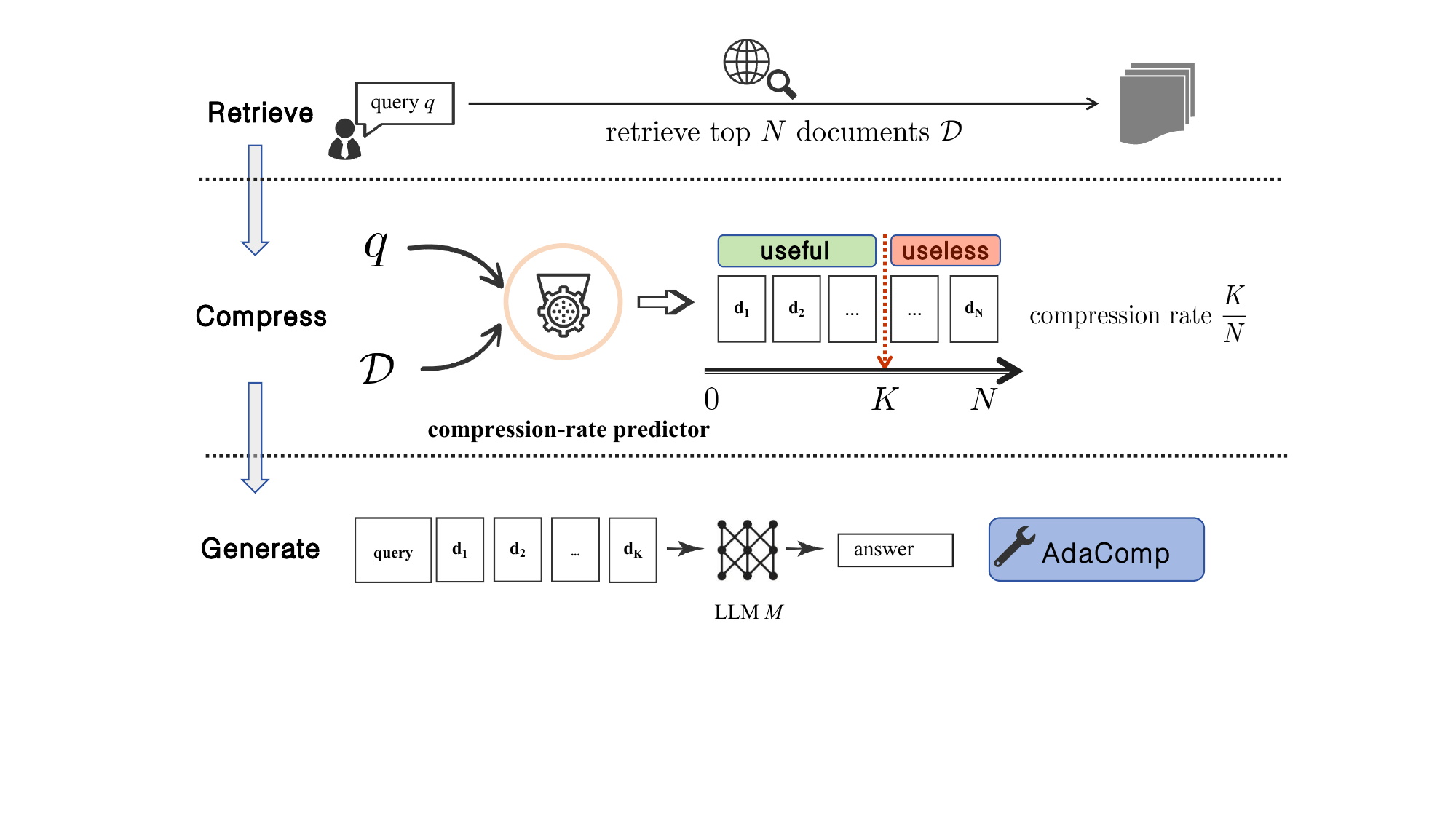} 
\caption{Overall architecture of AdaComp, which includes a retriever module \( R \), a compression module \( C_\theta \), and a generation module \( G \).}
\label{fig2}
\end{figure*}
\section{Related Work}

Given the inherent limitations of retrievers, the content they retrieve often contains noise, which can significantly undermine the accuracy of the generated output. Moreover, when the context provided to the model is excessively long, it can further diminish the model’s efficiency. To address these challenges, \citet{xu2023recomp} propose leveraging high-performance large language models (LLMs) to train summarization compressors that condense the retrieved texts. However, this approach is not without its flaws, as the generated summaries sometimes fail to faithfully represent the original content. Similarly, \citet{xu2023recomp,wang2023learning} explore the use of extractive compressors that identify and select the sentences most relevant to the query. While this method helps in filtering out irrelevant information, it also faces the risk of excessive compression, which can lead to a reduction in output accuracy. In a different approach, \citet{li2023compressing} introduce the concept of Selective Context, which aims to enhance the efficiency of LLMs during inference by eliminating redundant content based on self-information metrics. However, this technique may inadvertently disrupt the semantic coherence of the context. Finally, \citet{zhu2024information} apply the information bottleneck principle to filter noise, striving to strike a balance between conciseness and correctness. Despite its potential benefits, this method is associated with high computational complexity during the training process, posing additional challenges for practical implementation.
Several general compression methods have not been specifically tailored or optimized for Retrieval-Augmented Generation (RAG). For instance, \citet{ge2023context} and \citet{chevalier2023adapting} focus on compressing long contexts into short, compact memory slots that can be directly utilized by the large language model (LLM) for various downstream tasks. Additionally, \citet{jiang2023llmlingua,jiang2023longllmlingua} introduce a coarse-to-fine prompt compression technique based on the perplexity score, aiming to efficiently reduce the input size while maintaining the quality of the generated output. However, while these methods are more general-purpose, they have not been specifically optimized for the challenges and requirements of RAG systems.

\section{Method}
In this section, we first introduce the architecture of AdaComp, then describe how to obtain compression rate labels, train a compression-rate predictor, and integrate the compression model with generation models for inference, respectively.

\subsection{Model Architecture}
The Adaptive Context Compression Architecture is a novel framework designed to optimize document selection for retrieval-augmented generation tasks by balancing query complexity and retrieval quality. This architecture is composed of three main components: a retrieval module \( R \), a compression module \( C_\theta \), and a generation model \( G \), as shown in Figure~\ref{fig2}. 
Given a query $q$, a target response $y$, and a set of retrieved documents $\mathcal{D} = \{d_1, d_2, \dots, d_N\}$, unlike traditional approaches that handle all the retrieved documents indiscriminately, our method dynamically selects an optimal subset of documents \( \mathcal{D}' \subseteq \mathcal{D} \) that allows the generative model \( G \) to produce the target response $y$.
The compression module \( C_\theta \) is specifically trained to perform this document selection. It receives as input the query \( q \) and the full set of retrieved documents \( \mathcal{D} \) and outputs a compressed subset \( \mathcal{D}' \). This process reduces the computational burden on the generation model by limiting the input size without compromising the quality of the response. The main component of compression module \( C_\theta \) is a compression-rate predictor \( f(\mathcal{D}, q) \) which determines the size of \( \mathcal{D}' \) based on the query $q$ and the retrieved documents \( \mathcal{D} \). When the query is simple or the retrieval quality is high, the compressor selects fewer documents; conversely, for complex queries or lower-quality retrievals, it selects more documents. This ensures that the generation model \( G \) is provided with the most pertinent information, enhancing the accuracy and efficiency of the generated output. Therefore, our work focuses on how to train an efficient and low-cost compression-rate predictor.

\subsection{Predictor's Training Data}
To determine the optimal document subset \(\mathcal{D'}\) for each query \(q\), we employ a data annotation method based on real RAG system feedback. Given a query \(q\) and a set of retrieved documents \(\mathcal{D} = \{d_1, d_2, \dots, d_n\}\), the objective is to identify the smallest subset \(\mathcal{D'}\) such that the RAG system \(M\) can generate the correct response \(y\) based on \(q\) and \(\mathcal{D'}\).

Let $\mathcal{D}_k = \{d_1, d_2, \dots, d_k\}$ represent a subset of \(\mathcal{D}\) containing the top-k documents, where \(1 \leq k \leq n\). The system's performance is evaluated on each subset \(\mathcal{D}_k\) by checking if the system's output \(M(q, \mathcal{D}_k)\) matches the ground truth \(y\). The correctness condition is defined as:
\[
\textnormal{Correct}(q, \mathcal{D}_k) =
\left\{
\begin{array}{ll}
1, & \textnormal{if } M(q, \mathcal{D}_k) = y \\
0, & \textnormal{otherwise}
\end{array}.
\right.
\]
The process involves iterating over the subsets from the largest to the smallest, starting with \(\mathcal{D}_n\) and continuing to \(\mathcal{D}_1\), as shown in Algorithm~\ref{alg:1}. The optimal subset \(\mathcal{D'}\) is the smallest subset \(\mathcal{D}_k\) for which the system generates a correct response:

\[
\mathcal{D'} = \arg\min_{k} \left\{ k \mid \textnormal{Correct}(q, \mathcal{D}_k) = 1 \right\}.
\]

If the RAG system \(M\) can not generate a correct response for any subset, then \(\mathcal{D'} = \emptyset\), indicating that no subset of the retrieved documents suffices to produce the correct answer. This method ensures that the minimal necessary context \(\mathcal{D'}\) is used, thereby optimizing the balance between information relevance and computational efficiency.

\begin{algorithm}
\caption{Find the Optimal Subset $\mathcal{D}_k$}
\begin{algorithmic}[1]
\Require Query $q$, Document Set $\mathcal{D} = \{d_1, d_2, \dots, d_n\}$, Target Output $y$
\Ensure Optimal Subset $\mathcal{D}_k$ such that $M(q, \mathcal{D}_k) = y$
\State $min\_k \gets n + 1$ \Comment{Initialize minimum subset size}
\For{$k \gets 1$ to $n$}
    \State $\mathcal{D}_k \gets \text{SelectTopKDocuments}(\mathcal{D}, k)$
    \If{$M(q, \mathcal{D}_k) = y$}
        \State $min\_k \gets k$
        \State \textbf{break} \Comment{Stop when the smallest $k$ is found}
    \EndIf
\EndFor
\If{$min\_k \leq n$}
    \State \Return $\text{SelectTopKDocuments}(\mathcal{D}, min\_k)$
\Else
    \State \Return \textbf{None} \Comment{No valid subset found}
\EndIf
\label{algorithm}
\end{algorithmic}
\end{algorithm}

\begin{algorithm}[tb]
    \caption{Select Top K Documents}
    \label{alg:SelectTopKDocuments}
    \renewcommand{\algorithmicrequire}{\textbf{Input:}}
    \renewcommand{\algorithmicensure}{\textbf{Output:}}
    
    \begin{algorithmic}[1]
        \Require Document Set $\mathcal{D}$, Number of Documents $k$
        \Ensure Top $k$ Documents $\mathcal{D}_k$
        
        \Function{SelectTopKDocuments}{$\mathcal{D}, k$}
            \State $\mathcal{D}_k \gets \text{top } k \text{ documents from } \mathcal{D}$
            \State \Return $\mathcal{D}_k$ \Comment{Return the top $k$ documents}
        \EndFunction
        
    \end{algorithmic}
    
\end{algorithm}

\subsection{Compression-rate Predictor}
Given an input query \( q \) and the set of retrieved documents \( \mathcal{D} = \{d_1, d_2, \dots, d_N\} \), the goal of context compressor is to train a compression-rate predictor \( f(\mathcal{D}, q) \) to predict the number of documents \( |\mathcal{D}'| \) that should be selected from \( \mathcal{D} \) to generate a target response. We utilize the Llama2-7b model as the base model of the compression-rate predictor.
The compression-rate predictor \( f(\mathcal{D}, q) \) is trained to output a discrete number \( n \), where \( n \) can take any value from the set \(\{0, 1, \cdots, N\}\). The training process involves fine-tuning the Llama2-7b model on the above-annotated training dataset consisting of triples \((q, \mathcal{D}, \hat{n})\), where \( \hat{n} \) represents the true number of documents required.

The training objective is to minimize the classification loss between the predicted number of documents and the actual number required. Let \( \hat{n} \) be the true number of documents needed for optimal performance. The loss function \( \mathcal{L}(\theta) \) is defined as follows:

\[
\mathcal{L}(\theta) = - \sum_{i=1}^M \left[ \text{True}_i \cdot \log(\hat{p}_i) + (1 - \text{True}_i) \cdot \log(1 - \hat{p}_i) \right],
\]
where \( \text{True}_i \) is the true class label $\hat{n}$ for the \( i \)-th training example, and \( \hat{p}_i \) is the predicted probability of the \( i \)-th example belonging to each class.

During fine-tuning, the Llama2-7b model learns to map the input query \( q \) and document set \( \mathcal{D} \) to the appropriate class $\hat{n}$ representing the number of documents required. The optimization process updates the model parameters \( \theta \) to minimize the classification loss:

\[
\theta \leftarrow \theta - \eta \cdot \nabla_\theta \mathcal{L}(\theta),
\]

where \( \eta \) is the learning rate.

To evaluate the performance of the trained compressor, the predicted number of documents \( n \) is compared to the true number \( \hat{n} \). Metrics such as accuracy, precision, and recall are used to assess how effectively the compressor predicts the optimal number of documents.
Through this approach, we aim to train the compressor \( C_\theta \) to accurately determine the number of documents needed for generating high-quality responses based on the input query and retrieved documents.

\subsection{Utilizing Compressed Context for Generation}
We focus on leveraging a high-performance model to utilize compressed context effectively for generation tasks. This approach involves using a compressed subset of documents \( \mathcal{D}' \) to optimize computational efficiency while ensuring high-quality responses.

During training, for each query \( q \) and its associated oracle documents \( \mathcal{D}' \), the input to the generation model \( G \) is constructed by concatenating \( q \) with \( \mathcal{D}' \). The model is trained to produce the correct output \( o \) given this input, formalized as:
\[
G(o \mid q \oplus \mathcal{D}').
\]

For inference, a context \( \mathcal{D}_{\text{pred}} \) is derived from the full set of documents \( \mathcal{D} \). This filtered context $C_f$ is obtained by selecting the most relevant documents based on the model's compression-rate predictor. The input to the generation model during inference is:
\[
C_f = q \oplus \mathcal{D}_{\text{pred}},
\]
\[
\mathcal{D}_{\text{pred}} = \text{Top-K}( \mathcal{D}),
\]
where $\text{K}$ is determine with $f(\mathcal{D}, q)$.

\begin{table*}[t] 
    \centering
    \setlength{\tabcolsep}{3mm} 
    \begin{tabular}{lccccccccc}
        \toprule
         & \multicolumn{3}{c}{\textbf{NQ}} & \multicolumn{3}{c}{\textbf{TriviaQA}} & \multicolumn{3}{c}{\textbf{HotpotQA}} \\
        \textbf{Method}  & tokens & EM  & F1  & tokens & EM & F1  & tokens & EM  & F1 \\
        \midrule
        \textbf{\textit{No Retrieval}} \\
        LLAMA2-7B & - & 26.98 & 62.51 & - & 30.54 & 68.86 & - & 19.96 & 55.84 \\
       \midrule
        \textbf{\textit{Retrieval without Compression}} \\
        Top-1 document & 159 & 36.81 & 69.21 & 160 & 43.57 & 77.51 & 164 & 25.84 & 60.31 \\
        Top-5 documents & 802 & \textbf{40.64} & \textbf{71.09} & 808 & \textbf{48.58} & \textbf{80.20} & 819 & 25.09 & 59.56 \\
       \midrule
        \textbf{\textit{Retrieval with Compression}} \\
        RECOMP & 67 & 32.85 & 66.08 &74 & 41.77 & 76.17 & 103 & 24.59 & 59.23 \\
        FILCO & 46 & 32.43 & 64.78 & 58 & 38.96 & 74.14 & 76 & 20.12 & 56.03 \\
        ONLY\_DOC & 162 & 36.59 & 69.11 & 166 & 44.60  & 77.86 & 173  & 25.79 & 60.13 \\
       \midrule
         \textbf{\textit{Adaptive Compression}} \\
        Oracle & 262 & 52.11 & 76.28 & 261 & 59.36 & 83.98 & 348 & 33.71 & 64.28 \\
        \textbf{Ours} & 441 & \textbf{40.13} & \textbf{70.96} & 468 & \textbf{47.15} & \textbf{79.40} & 527 & \textbf{26.36} & \textbf{60.46} \\
        \bottomrule
    \end{tabular}   
    \caption{ Results(\%) on the three open-domain QA datasets with LLAMA2-7B as the generator.}
    \label{tab:table1}
\end{table*}

\section{Experiments}
In this section, we will introduce datasets, evaluation metrics, settings, baselines, and further analysis.
\subsection{Experimental Settings}

\subsubsection{Datasets} We evaluate our adaptive content compression method on three benchmark datasets: Natural Questions (NQ) \cite{kwiatkowski2019natural}, TriviaQA \cite{joshi2017triviaqa}, HotpotQA \cite{yang2018hotpotqa} and conversational Multi-doc QA~\footnote{https://sites.google.com/view/wsdm24-docqa}. We utilize the adversarial Dense Passage Retriever (DPR) \cite{karpukhin2020dense} to retrieve the Top-5 passages from the full Wikipedia passages for each QA dataset.

\subsubsection{Evaluation Metrics} For the open-domain QA datasets, we assess end-task performance using Exact Match (EM) and F1 Score for the answer strings following other work~\cite{wang2025maferw,chen2025privacy}. EM assesses exact correctness, while F1 evaluates answers that are close to but not necessarily exact, providing a nuanced view of how well-predicted answers overlap with the correct ones. For the conversational Multi-Doc QA dataset, we use Rouge-1, Rouge-2, and Rouge-L to evaluate the quality of the generated responses. Each metric represents performance at a different level: Rouge-1 assesses unigram (single word) overlap, Rouge-2 measures bigram overlap, and Rouge-L evaluates the longest common subsequence overlap, reflecting sentence-level coherence and fluency.

\begin{table}[t] 
    \centering
    \setlength{\tabcolsep}{1mm} 
    \begin{tabular}{lcccc}
        \toprule
        \textbf{Dataset} & \textbf{Method} & \textbf{Rouge-1} & \textbf{Rouge-2} & \textbf{Rouge-L} \\
        \midrule
        specific & Top-1 & 43.22 & 18.05 & 39.71\\
                   & Top-5 & 44.94 & 18.84 & 41.29 \\
                   & Top-Random & 44.89& 18.97 & 41.24\\
                   & RECOMP & 38.41 & 15.52 & 35.36\\
                    & Oracle & 51.66  & 24.14& 47.72\\
                  &  Ours & \textbf{46.89} &  \textbf{20.56}& \textbf{43.19}\\
         
        \midrule

         open-ended & Top-1 & 44.01 & 19.23 & 40.37 \\
                   & Top-5 & 46.25 & 20.52 & 42.66\\
                   & Top-Random &46.02  & 20.51 &42.24 \\
                    & RECOMP & 38.82 &15.08  & 35.40\\    
                   & Oracle & 53.94  & 26.79 & 49.98\\
                  &  Ours & \textbf{48.75} &  \textbf{22.67} & \textbf{45.01}\\

       \bottomrule
    \end{tabular}
    \caption{Results(\%) on the conversational Multi-doc QA dataset with LLAMA2-7B as the predictor and LLAMA2-13B-chat as the generator.}
    \label{tab:table2}
\end{table}
\subsubsection{Settings}
We use LLAMA2~\cite{touvron2023llama} as the backbone architecture for the large language model. We finetune the 7B model version with LORA~\cite{hu2021lora} as the compression-rate predictor for 10 epochs on a single 40G NVIDIA A100, using approximately 18 hours. The initial learning rate is set to 5e-4, batch size is set to 8, and the proportion of warmup steps is set to 0.1. We select the best model based on the performance of the validation set. In the subsequent generation phase, we utilize the LLAMA2-7B model for three open-domain QA datasets and the LLAMA2-13B-Chat model for the conversational Multi-Doc QA dataset.

\subsubsection{Baselines} We select four types of baselines, including no retrieval, retrieval without/with compression, and oracle.

\textbf{No Retrieval:} This baseline represents the scenario where no external documents are retrieved or provided to the model. The model generates responses solely based on its internal knowledge without any retrieval augmentation. This approach serves as a baseline to demonstrate the benefits of retrieval methods in enhancing response quality.

\textbf{Retrieval without Compression:} In this setup, the model utilizes the top-k retrieved documents without applying any form of compression. We experiment with two configurations: Top-1 document and Top-5 documents. The goal is to assess the impact of using uncompressed retrieval augmentation. We also use the Top-Random baseline method, where documents are randomly selected from the retrieved documents to assess performance across different numbers of top results.

\begin{figure*}[t]
\centering
\includegraphics[width=0.8\textwidth]{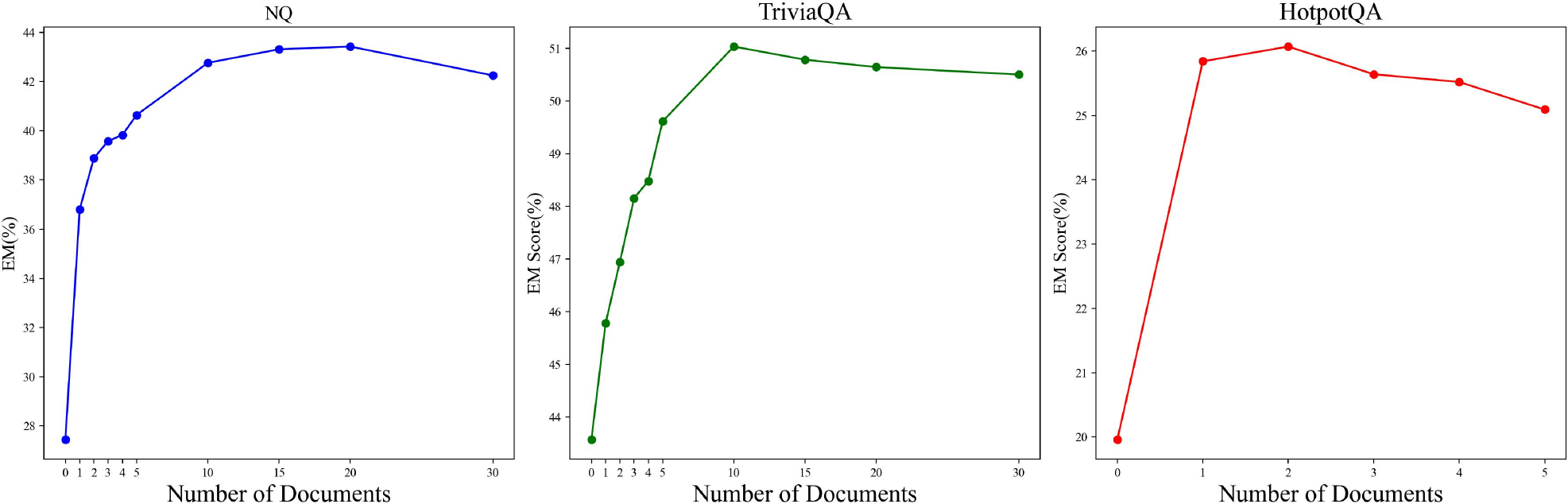} 
\caption{An illustration of how the number of documents affects final RAG performance, generally, in the beginning, as the number of documents increases, RAG performance improves due to the provision of sufficient information. However, as the number of documents increases excessively, the inclusion of a large amount of noise leads to a decline in RAG performance.}
\label{fig:figure3}
\end{figure*}
\textbf{Retrieval with Compression:} 
We select two baseline compression methods: RECOMP for extractive compression and FILCO for abstractive compression. RECOMP filters retrieved documents by using a trained extractor to select key sentences directly relevant to the query, providing a focused subset to the generation model. FILCO, on the other hand, generates summaries from relevant sentences using a summarization model. We also experimented with a compression method that uses only the document containing the sentence most relevant to the query as context, referred to as ONLY\_DOC.

\subsubsection{Oracle} The Oracle setting represents a theoretical upper bound for performance, where the most relevant documents and the optimal compression rates are provided to the model. In this scenario, the selection of documents and the degree of compression are perfectly aligned with the model’s requirements, leading to the best possible outcomes. The Oracle serves as a reference point to compare the effectiveness of the proposed methods and baselines, highlighting the potential performance gains that could be achieved with an ideal retrieval and compression strategy.

\subsection{Main Results}
We report the results on three QA datasets in Table~\ref{tab:table1}. We find that all retrieval augmentation methods improve performance over the no retrieval setting, indicating that retrieval operations are crucial for enhancing model performance. Secondly, by retrieving documents without compression (e.g., Top 1 and Top 5 documents), the model shows significant improvements in EM and F1 scores across all datasets, with a further boost in performance when using the Top 5 documents. This suggests that retrieving more relevant documents can improve the quality of the model’s answers. The Oracle method (theoretical best) outperforms all other methods, particularly on the NQ and TriviaQA datasets, demonstrating that adaptively selecting the appropriate compression rate can greatly enhance model performance. Our AdaComp method performs better across all three datasets compared to the RECOMP and FILCO methods. It also significantly reduces the number of tokens, indicating that the adaptive compression strategy strikes a good balance between token count and performance. The RECOMP and FILCO methods, by compressing the retrieved documents, significantly reduce the number of tokens but show significantly lower performance in terms of EM and F1 compared to the uncompressed retrieval methods. This suggests that while compression strategies effectively reduce input size, they may lead to performance degradation due to over-compression.

We report the results on one conversational dataset in Table~\ref{tab:table2}. We divide the test dataset into specific and open-ended questions according to the following rule: we calculate the top 5 relevance scores between each document and the given answer. If the maximum relevance difference exceeds 0.3, the query is classified as a specific question; otherwise, as an open-ended question. From the results of Table~\ref{tab:table2}, we can see that our AdaComp method outperforms the baseline on both specific and open-ended questions, with a more pronounced advantage on open-ended questions, demonstrating that our compression approach effectively recognizes query complexity and provides an optimal compression rate.

\begin{table}[t] %
    \centering
   \setlength{\tabcolsep}{1mm} 
    \begin{tabular}{lcccc}
        \toprule
        \textbf{Dataset} & \textbf{Method} & \textbf{EM} & \textbf{F1} & \textbf{Avg. docs} \\
        \midrule
          NQ & Top-Random & 39.26 & 70.22 & 3.00\\
          & Top-2 & 38.89 & 70.16 & 2.00\\
          & Top-3 & 39.58 & 70.53 & 3.00\\
             & Top-4 & 39.83 & 70.76 & 4.00\\
        &  Ours & \textbf{40.13} & \textbf{70.96} & 3.66 \\
        \midrule
       TriviaQA & Top-Random & 46.72 & 79.28 & 2.99 \\
                 & Top-2 & 45.78 & 78.91 & 2.00 \\
                & Top-3 & 46.85 & 79.64 & 3.00 \\
                 & Top-4 & \textbf{48.15} & \textbf{79.92} & 4.00 \\
        & Ours & 47.15 & 79.40 & 3.23 \\
        \midrule
        HotpotQA & Top-Random & 25.54 & 60.11 & 2.97 \\
          & Top-2 & 26.07 & 60.23 & 2.00 \\
          & Top-3 & 25.64 & 60.14 & 3.00 \\
          & Top-4 & 25.52 & 60.01 & 4.00 \\
        &  Ours  & \textbf{26.36} & \textbf{60.46} & 2.13\\
       \bottomrule
    \end{tabular}
  
    \caption{Results(\%) on AdaComp compared to Top-Random method.}
    \label{tab:table3}

\end{table}

\subsection{The Impact of Document Number} \label{sec:moExam}

As illustrated in Figure~\ref{fig:figure3}, the number of documents utilized in RAG significantly affects performance. Initially, performance improves with an increasing number of documents, as the inclusion of additional relevant information enhances the model’s accuracy and the breadth of its responses. This initial improvement, however, is not without limits. Beyond a certain threshold, further increases in the number of documents lead to a decline in performance. This decline is primarily due to the influx of noisy or irrelevant data, which dilutes the quality of the retrieved information and impairs the model’s ability to effectively filter out noise. Consequently, the model struggles to maintain high-performance levels as the proportion of valuable information decreases. Thus, optimizing the number of documents is crucial for balancing information and noise, which is essential for maximizing RAG performance.

\subsection{Effectiveness of the Predictor}

Based on the results presented in Table 3, our method demonstrates robust performance across a range of metrics, showcasing its capability to achieve high effectiveness while maintaining a lower average document count. Specifically, for the NQ dataset, our method achieves an Exact Match (EM) score of 40.13\% and an F1 score of 70.96\%. This performance not only surpasses that of the Top-4 method but also does so while keeping a lower average document count of 3.66, illustrating the efficiency and precision of our approach in handling the given data. In the TriviaQA dataset, although the Top-4 method slightly exceeds our approach in terms of EM and F1 scores, with figures of 48.15\% and 79.9\% respectively, our method still achieves commendable results. We attain an EM score of 47.15\% and an F1 score of 79.40\% with a reduced average document count of 3.23. This demonstrates a notable balance between performance and document efficiency, where our approach manages to deliver high-quality results with fewer documents. On the HotpotQA dataset, our method stands out by outperforming all other methods. We achieve an EM score of 26.36\% and an F1 score of 60.46\% while maintaining an exceptionally low average document count of just 2.13.

\begin{figure}[!t]
\centering
\includegraphics[width=0.9\columnwidth] {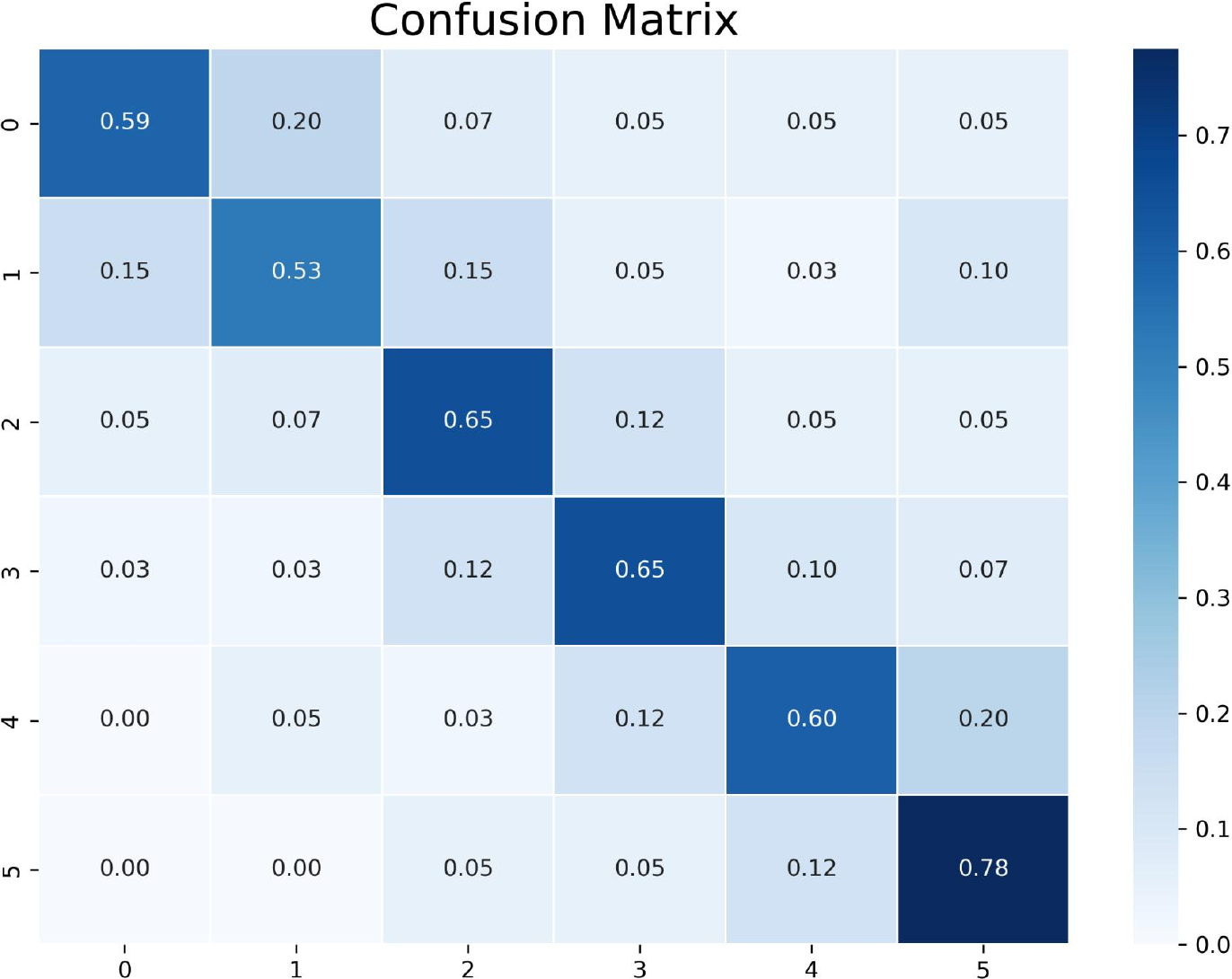} 
\caption{Confusion Matrix for Predictor Performance.}
\label{fig:figure4}
\end{figure}

\subsection{Performance of the Predictor}

The confusion matrix for our predictor’s performance is illustrated in Figure 4. Our approach classifies the compression rate into six distinct categories, providing a detailed view of how effectively our model differentiates between various levels of compression. The overall accuracy of our predictions is approximately 65\%, indicating that AdaComp demonstrates a reasonably good predictive capability. Additionally, an analysis of the confusion matrix reveals that the absolute difference between the predicted labels and the true labels typically falls within a margin of 2. This finding suggests that our predictor is quite effective in estimating the required compression range. The relatively small discrepancy indicates that, while the model may not always predict the exact compression rate, it remains consistently close to the actual required values. This demonstrates the model’s reliable performance in estimating the necessary level of document compression.
\subsection{Case Study}

\begin{figure}[!t]
\centering
\includegraphics[width=0.9\columnwidth] {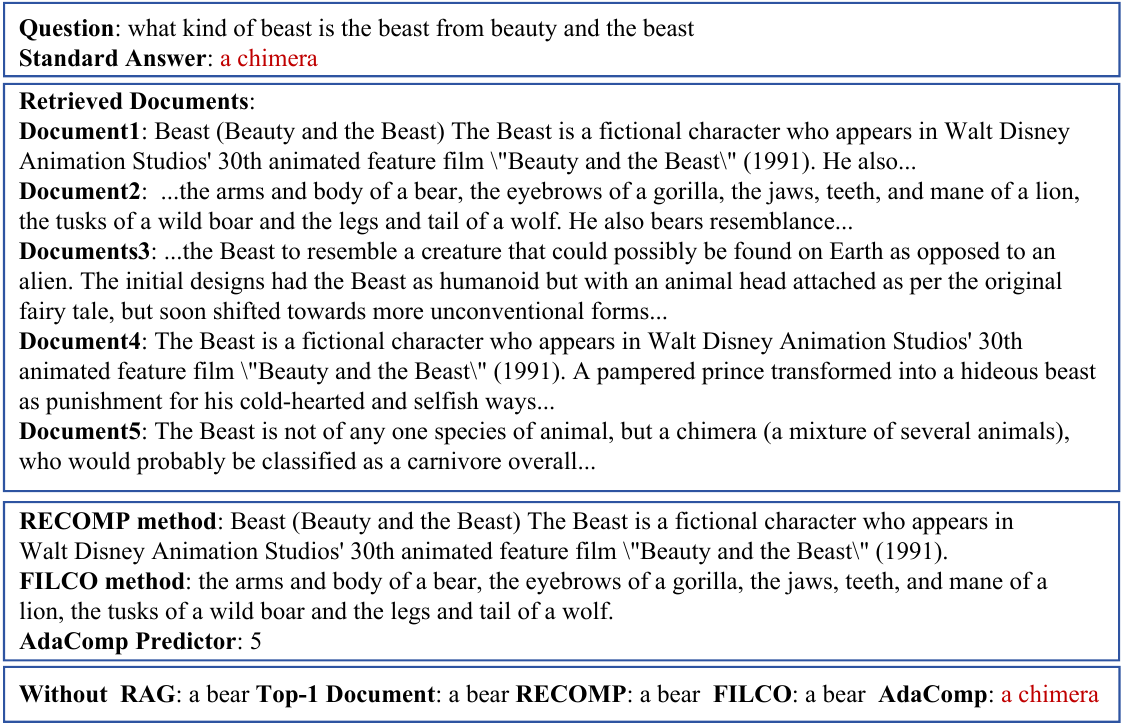} 
\caption{Case Study: answers generated using without RAG, Top-1 document, RECOMP, FILCO, and AdaComp.}
\label{fig:figure5}
\end{figure}

As shown in Figure~\ref{fig:figure5}, AdaComp demonstrates the ability to retain a greater amount of relevant information when the quality of the retrieved text is suboptimal. This enhanced retention capability allows AdaComp to generate accurate answers despite the lower quality of the input text. In contrast, other compression methods struggle to produce correct responses under similar conditions. This performance discrepancy highlights AdaComp’s robustness in handling less-than-ideal retrieval scenarios, ensuring that the quality of the generated answers is maintained even when the initial text quality is compromised.

\section{Conclusion}

This paper introduces a low-cost but effective context compression method, AdaComp, which adaptively determines the compression rate based on both query complexity and retrieval quality. When handling complex questions or low-quality retrieved documents, AdaComp retains more context to ensure the final performance of RAG. Conversely, when dealing with simpler questions or high-quality retrieved documents, AdaComp adaptively compresses the context to be both sufficient and concise, thereby enhancing RAG’s compression efficiency.
In future work, we will investigate whether the length of retrieved documents influences LLMs’ ability to answer questions, with a focus on the impact of context length. Additionally, we will explore new methods to more finely distinguish situations where the required number of documents is close, aiming to improve the accuracy of the final predictor.

\bibliography{aaai25}

\end{document}